%% file: main.tex
\documentclass[10pt,twocolumn,letterpaper]{article}

\usepackage{cvpr}              
\input{preamble}
\definecolor{cvprblue}{rgb}{0.21,0.49,0.74}
\usepackage[pagebackref,breaklinks,colorlinks,allcolors=cvprblue]{hyperref}
\usepackage{adjustbox}
\usepackage{algorithm}
\usepackage{amsmath,amssymb,mathtools}
\usepackage{graphicx}
\usepackage{algpseudocode}
\usepackage{float}

\def\paperID{1491} 
\def\confName{CVPR}
\def\confYear{2026}

\title{Where, What, Why: Toward Explainable 3D-GS Watermarking}
\author{
Mingshu Cai$^{1}$ \quad
Jiajun Li$^{2}$ \quad
Osamu Yoshie$^{1}$ \quad
Yuya Ieiri$^{1}$ \quad
Yixuan Li$^{3}$\thanks{Corresponding author.}
\\
$^{1}$Waseda University \quad
$^{2}$Southeast University \quad
$^{3}$Nanyang Technological University
\\
{\tt\small mignshucai@fuji.waseda.jp, yoshie@waseda.jp, ieyuharu@ruri.waseda.jp}
\\
{\tt\small jiajun\_li@seu.edu.cn, yixuan.li@ntu.edu.sg}
}

\begin{document}
\maketitle
\input{sec/0_abstract}    
\input{sec/1_intro}

\input{sec/2_related_work}

\input{sec/3_method}
\input{sec/4_experiment}

\input{sec/5_conclusion}
{
    \small
    \bibliographystyle{ieeenat_fullname}
    \bibliography{main}
}


\end{document}

%% file: sec/0_abstract.tex
\begin{abstract}
As 3D Gaussian Splatting becomes the de facto representation for interactive 3D assets, robust yet imperceptible watermarking is critical. We present a representation-native framework that separates where to write from how to preserve quality. 
A Trio-Experts module operates directly on Gaussian primitives to derive priors for carrier selection, while a Safety and Budget Aware Gate (SBAG) allocates Gaussians to watermark carriers—optimized for bit resilience under perturbation and bitrate budgets—and to visual compensators that are insulated from watermark loss. 
To maintain fidelity, we introduce a channel-wise group mask that controls gradient propagation for carriers and compensators, thereby limiting Gaussian parameter updates, repairing local artifacts, and preserving high-frequency details without increasing runtime. Our design yields view-consistent watermark persistence and strong robustness against common image distortions such as compression and noise, while achieving a favorable robustness–quality trade-off compared with prior methods. 
In addition, the decoupled finetuning provides per-Gaussian attributions that reveal where the message is carried and why those carriers are selected, enabling auditable explainability. Compared with state-of-the-art methods, our approach achieves a PSNR improvement of +0.83 dB and a bit-accuracy gain of +1.24\%.

\end{abstract}

%% file: sec/1_intro.tex
\section{Introduction}
\label{sec:intro}

3D Gaussian Splatting (3D-GS) \cite{3D-GS}, with its explicit parameterization, real-time performance, and high fidelity, is emerging as the mainstream paradigm for 3D content creation and deployment—succeeding NeRF~\cite{nerf}—and is being widely adopted across film, gaming, autonomous driving, digital humans, and world models \cite{hugs,gs_world,gbc-splat,langsplat,langsplatv2,splatad}.

\begin{figure}[t]
    \centering
    \includegraphics[width=\linewidth]{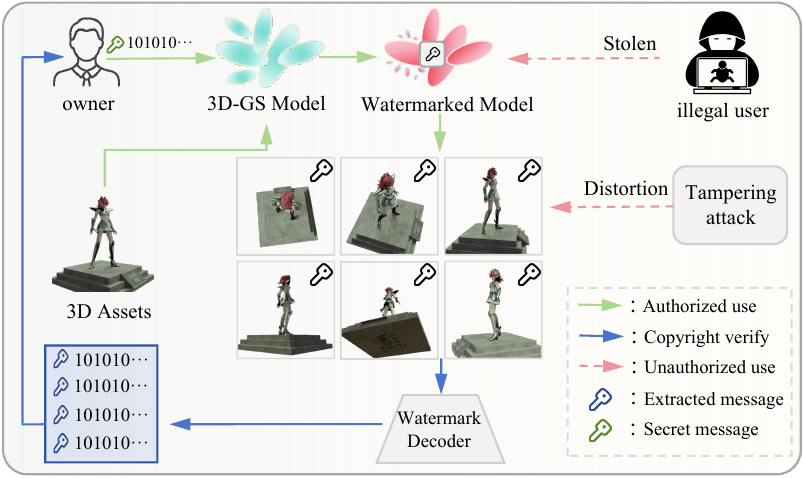}
    \caption{\textbf{Usage of our model.}
    The owner embeds a secret message into a 3D-GS model using our method; even if the model is stolen and undergoes distortion attacks, the rendered views can still be decoded to verify copyright.}
    \label{fig:overview}
    \vspace{-1.4em}
\end{figure}

However, with the large-scale generation and widespread distribution of 3D assets, their susceptibility to copying and tampering has made copyright protection increasingly critical—once a model is illicitly obtained or redistributed, the original creator’s rights become extremely difficult to enforce.
This challenge is particularly acute in 3D Gaussian Splatting (3D-GS): its core strength—explicit, directly editable Gaussian parameters—while enhancing representational expressiveness and rendering efficiency, also introduces significant security risks.
Attackers can easily copy the model, tamper with its content, strip away authorship information, and illegally redistribute it, thereby severely undermining copyright attribution and provenance tracking.

Effective watermarking for 3D-GS is therefore urgent. Prior work on radiance-field watermarking, such as WateRF and 3DGSW \cite{waterf,3dgsw,guardsplat}, leaves two core gaps for explicit, discretized Gaussians:

(i) Carrier selection: From a large, heterogeneous set of Gaussian primitives, select watermark carriers by jointly considering multi-view visibility, frequency-domain cues, and the stability of geometry and appearance, with an emphasis on visual stability and information security.

(ii) Robust \& imperceptible embedding: Embed a robust watermark without degrading visual or rendering quality, and ensure it remains extractable after common distortions such as cropping, compression, and format conversion.

Motivated by these considerations, we aim to build a framework unifying where to write, what to write, and why it matters: selecting stable 3D Gaussian carriers (where), embedding a distortion-resilient yet invisible signal (what), and achieving interpretable, auditable watermarking (why). A brief usage example is shown in Fig.~\ref{fig:overview}

To address carrier selection, we introduce Trio-Experts, which analyzes intrinsic 3D parameters rather than rendered images. It extracts representation aligned priors: the Geometry expert scores structural stability, the Appearance expert gauges frequency characteristics for imperceptible edits, and the Redundancy expert estimates spatial replaceability for robustness. An uncertainty-aware fusion yields robustness-optimized Gaussian priors. These priors feed the Safety and Budget Aware Gate (SBAG), which finalizes \textbf{where to write} by routing Gaussians to the watermark set (WM) only when multi-view stability and safety budgets are met, and by expanding/densifying WM within the permitted visual-quality envelope. Non-carriers are assigned as Visual Compensators (VIS) for watermark embedding.

To minimize visual degradation from both WM and VIS Gaussians and improve embedding efficiency, we introduce a channel-wise Group Mask that specifies \textbf{what to write} by selecting watermark-eligible parameter channels while constraining gradient flow and per-channel update magnitudes.

Crucially, we completely decouple WM and VIS during training. VIS points are excluded from the watermark loss to avoid conflicts between carrier optimization and rendering fidelity, which explains \textbf{why} the system remains stable under adversarial procedures such as EOT; if VIS were coupled they would counter WM updates, harming visual quality and destabilizing extraction. Decoupling preserves image quality while maintaining watermark accuracy. In essence, this dual-role architecture enables robust \& imperceptible watermark embedding.

Extensive experiments show that our method embeds watermarks consistently across all rendered views of a 3D-GS model while remaining robust to attacks on both images and the underlying representation. Compared with state-of-the-art approaches \cite{waterf,guardsplat,3dgsw}, it achieves superior results across all major metrics. Our core contributions are summarized as follows:
\begin{itemize}
  \item We propose a highly interpretable, attack oriented \textbf{Decoupled Finetuning} framework that fully separates carriers and compensators updating, with a \textbf{channel wise Group Mask} to route WM and VIS gradients, suppress harmful changes, preserve high frequency details, and achieve secure, robust embedding.
  \item We introduce \textbf{Trio-Experts}, which operate directly on large 3D-GS point sets to extract high-quality geometry/appearance/redundancy priors for densification, channel-mask construction and watermark-carrier selection.
  \item We present the \textbf{Safety and Budget-Aware Gate (SBAG) } that, under an adaptive budget, uses representation priors and lightweight rendering to select and densify watermark carriers, while cleanly separating them from visual compensators to enable decoupled optimization.
  \item Our method achieves \textbf{state-of-the-art performance} and remains robust under diverse image distortion attacks.
\end{itemize}

%% file: sec/2_related_work.tex
\section{Related Work}
\label{sec:related work}

\begin{figure*}[ht!]
    \begin{center}
            \includegraphics[width=1\textwidth]{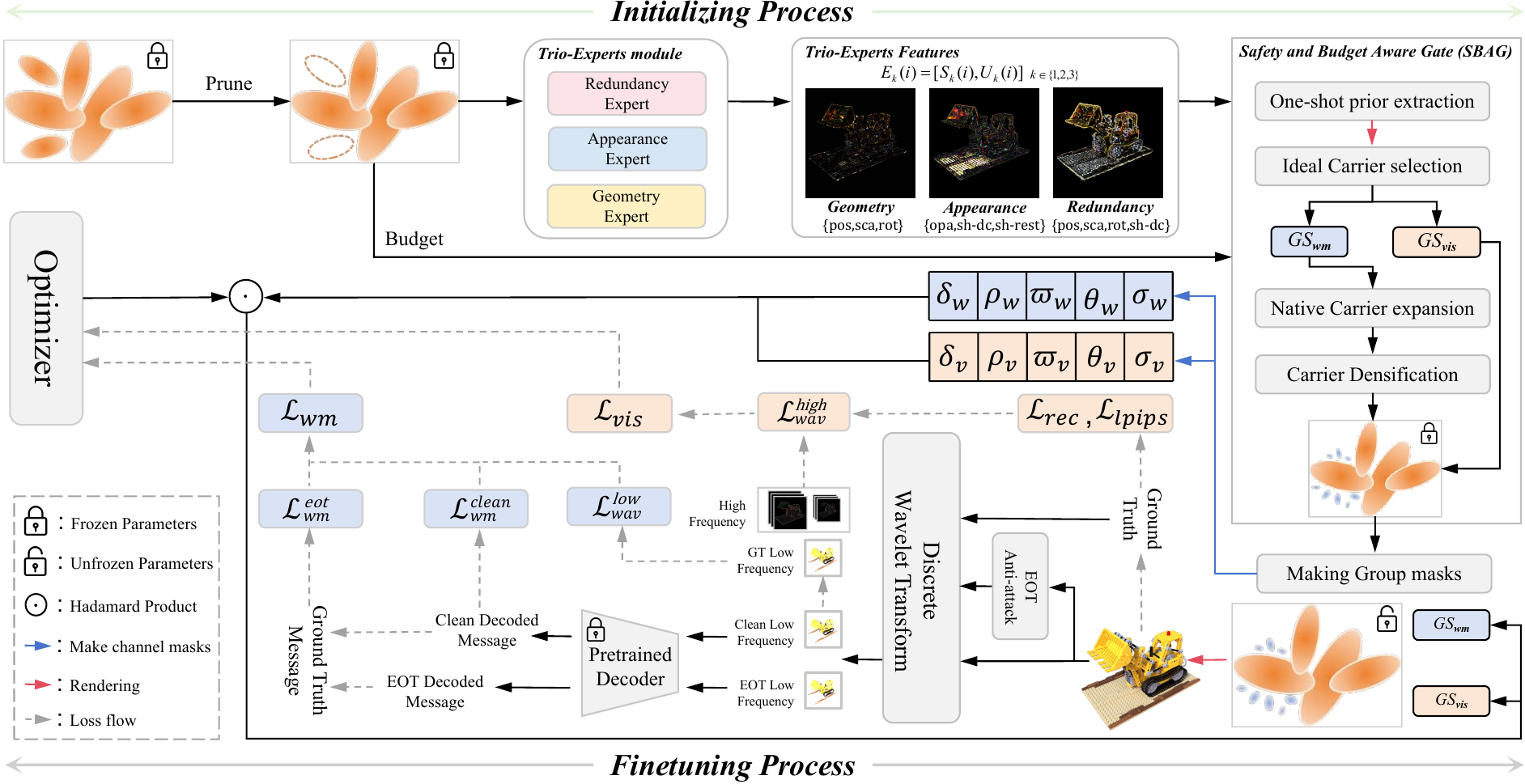}
    \end{center}
    \vspace{-1.5em}  
    \caption{\textbf{Pipeline of Our Method.} During initialization, we prune redundant Gaussians based on their rendering contribution. The \textbf{Trio-Experts} module extracts geometry/appearance/redundancy priors and aggregates them into an evidence package $E_k(i)$. The \textbf{SBAG} decouples ranking and budgeting and uses this evidence and a one-shot render to select, expand, and densify watermark carriers. In finetuning, a channel-wise \textbf{group mask} enforces disjoint gradient routing for watermark carriers ($GS_{\text{wm}}$) and visual compensators ($GS_{\text{vis}}$). EOT attacks render both clean and attacked views: $\mathcal{L}_{\text{vis}}$ preserves appearance, while low-frequency subbands form the watermark loss $\mathcal{L}_{\text{wm}}$, with $\mathcal{L}_{\mathrm{wav}}^{\mathrm{low}}$ penalizing over-editing. The separate optimization of $\mathcal{L}_{\text{vis}}$ and $\mathcal{L}_{\text{wm}}$ improves both fidelity and robustness.
}
    \label{fig:pipeline}
    \vspace{-1em}  
\end{figure*}

\subsection{3D Gaussian Splatting}
3D Gaussian Splatting (3D-GS) \cite{3D-GS} pairs an explicit point representation with differentiable splatting for real-time neural radiance-field rendering. Recent progress spans static reconstruction—shading/reflectance, large-scale scenes, structured/view-adaptive models, multi-scale anti-aliasing \cite{gshader,vastgaussian,imaggarment,msaa}—and dynamic/editable settings, including sparsely controllable editing, geometry-prior deformation, real-time 4D, and pose-conditioned human avatars \cite{sc-gs,imagpose,4DGS,AnimatableGS}. On the systems side, volumetrically consistent rasterization, 4K-scale efficiency, and training-time point dropping enhance physical fidelity, throughput, and regularization \cite{Vol3DGS,imagdressing,dropGS}; under weak constraints, depth–GS coupling and unposed joint estimation enable feed-forward reconstruction and cross-scene generalization \cite{depthsplat,flare,citygs-x}, while 3D-GS also probes 2D foundation features and unifies 2D/3D tone mapping for HDR \cite{feat2gs,gausshdr}. These trends sharpen the need for copyright protection and trustworthy use of 3D-GS assets.

\subsection{Discrete Wavelet Transform}
Discrete Wavelet Transform (DWT) captures spatial–frequency locality via multiscale decomposition and has long supported image denoising/detail recovery \cite{DWT}. In NeRFs and related 3D pipelines, wavelet-space priors improve sparsity, generalization, and training stability—covering coefficient sparsification and generalizable synthesis \cite{masked_dwt,wavenerf_dwt}, direction-aware dynamics \cite{darenerf_dwt}, and sparse-view wavelet losses plus 3D-GS frequency regularization \cite{dwtnerf,dwtgs,3dgsw}. We therefore enforce multiscale and local high-frequency consistency in the wavelet domain to boost quality and robustness while preserving global structure.
\subsection{Digital Watermarking}
Digital watermarking protects digital assets by identifying copyrights. The main difference lies in the priority of data embedding: watermarking prioritizes robustness—ensuring detection after distortions—whereas steganography prioritizes invisibility. To achieve robustness, traditional methods embed data in DWT subbands \cite{pixel-wise_marking,dwt_wm,dwt_wedding,robust_wm}. HiDDeN \cite{hidden} introduces an end-to-end deep watermarking framework with a noise layer. For radiance fields, CopyRNeRF \cite{copyrnerf} embeds messages into images rendered from implicit NeRFs. StegaNeRF \cite{steganerf} focuses on hiding information within rendered views with minimal visual impact, emphasizing imperceptibility. NeRFProtector \cite{nerfprotector} approaches the problem from a protection and authentication perspective, adding verifiable information to NeRFs without altering scene structure. WateRF \cite{waterf}, on the other hand, leverages DWT-based frequency embedding to enhance both fidelity and robustness. In the 3D-GS representation, 3DGSW \cite{3dgsw} improves robustness through joint regularization based on rendering contribution and wavelet-domain constraints, while GuardSplat \cite{guardsplat} focuses on real-world attack scenarios, achieving robust watermark embedding and verification with a strong pretrained CLIP \cite{clip} decoder.
Unlike prior frequency-cue methods, we watermark directly in 3D-GS parameter space with gated, decoupled, attack-aware training for secure, robust, visually lossless embedding.


%% file: sec/3_method.tex

\section{Method}
\label{sec:method}

\subsection{Preliminary}

\textbf{3D Gaussian Splatting.} We work under the standard 3D Gaussian Splatting (3D-GS) formulation~\cite{3D-GS}, where a scene is represented by a set of 3D Gaussian primitives parameterized by mean $\boldsymbol{\mu}$ and covariance $\boldsymbol{\Sigma}$:
\begin{equation}
G(\mathbf{x}; \boldsymbol{\mu}, \boldsymbol{\Sigma})
= \exp\!\left(-\tfrac{1}{2}\,(\mathbf{x}-\boldsymbol{\mu})^{\top}\boldsymbol{\Sigma}^{-1}(\mathbf{x}-\boldsymbol{\mu})\right).
\label{eq:gaussian3d}
\end{equation}
For rendering, each 3D Gaussian is projected to the image plane of a viewpoint $\pi$. Given depth ordering, pixels are composited by front-to-back alpha blending:
\begin{equation}
I_{\pi}[x,y] = \sum_{i \in \mathcal{N}_G} c_i \,\alpha_i \, T_i,
\quad
T_i = \prod_{j=1}^{i-1}\!\left(1-\alpha_j\right),
\label{eq:render}
\end{equation}
where $\mathcal{N}_G$ is the set of visible 2D Gaussians at $\pi$ sorted by depth, $c_i$ denotes color, $\alpha_i$ is the per-pixel contribution, and $T_i$ is accumulated transmittance.

\noindent\textbf{Prune by Contribution of Rendering Quality.} Pre-trained 3D-GS models often contain redundant 3D Gaussians. We adopt the contribution-based pruning method from 3D-GSW \cite{3dgsw}, which quantifies contribution using an auxiliary loss with temporary color parameters $C'$. The gradient $V_\pi = \partial L_\pi^{aux}/\partial C'$ serves as the contribution score to identify and prune low-impact Gaussians.

\subsection{3D Feature-Aware Trio-Experts}
\label{subsec:trio_experts}

Prior methods often select carrier locations in the image or frequency domain based on gradient or high-frequency heuristics, but this evidence tends to drift with viewpoint changes. To address this issue, we propose a representation-native Trio-Experts system, whose decision evidence is fully anchored in the 3D-GS parameter space. 

For the $i$-th Gaussian, its parameters are denoted as $\mathbf{x}_i \in \mathbb{R}^3$ (position), $\mathbf{s}_i \in \mathbb{R}_+^3$ (scale), $\mathbf{q}_i \in \mathbb{H}$ (rotation), $\alpha_i \in (0,1)$ (opacity), and $\mathbf{h}_i=[\mathbf{h}_i^{(0)}, \mathbf{h}_i^{(\ge 1)}]$ (SH coefficients). We only extract features from these native parameters, without relying on pixel-domain gradients, thereby ensuring view-consistent evaluation and reducing computational overhead. We group the native 3D-GS parameters by semantics:
$\mathcal{C}_{\text{geo}}=\{\mathbf{x},\mathbf{s},\mathbf{q}\}$,
$\mathcal{C}_{\text{app}}=\{\alpha,\mathbf{h}^{(0)},\mathbf{h}^{(\ge 1)}\}$,
and $\mathcal{C}_{\text{red}}=\{\mathbf{x},\mathbf{s},\mathbf{q},\mathbf{h}^{(0)}\}$.
After quantile-based min--max normalization, we construct a $k$-NN neighborhood $\mathcal{N}_k(i)$ in the 3D position space of $\mathbf{x}$ to model local density, and compute:\\

\noindent\textbf{Geometric Features $z_1$.}
Based on $\mathcal{C}_{\text{geo}}$, we capture structural decomposition and boundary cues in 3D space via scale isotropy, rotational consistency, and a compact footprint, to measure geometric stability:
\begin{equation}
z_1(i)=\mathrm{Norm}\big(\mathrm{Iso}_i,\ \mathrm{RotCons}_i,\ 1-\overline{\mathrm{fp}}_i\big),
\label{eq:z1}
\end{equation}
where $\mathrm{Iso}_i = \min(\mathbf{s}_i)/\max(\mathbf{s}_i)$ measures scale isotropy, $\mathrm{RotCons}_i = (1/k)\sum_{j\in\mathcal{N}_k(i)} |\langle \mathbf{q}_i,\mathbf{q}_j\rangle|$ is neighborhood quaternion consistency, and $\overline{\mathrm{fp}}_i = \exp((1/3)\sum_d\log s_{i,d})$ is the geometric-mean footprint.\\

\noindent\textbf{Appearance Features $z_2$.}
Based on $\mathcal{C}_{\text{app}}$, we measure cross-view appearance consistency from color and opacity cues via DC band-pass, opacity gating, and high-frequency suppression:
\begin{equation}
z_2(i)=\mathrm{Norm}\big(1-\rho^{\mathrm{hf}}_i,\ g(\alpha_i),\ c_i\big),
\label{eq:z2}
\end{equation}
where $\rho^{\mathrm{hf}}_i$ is the AC high-frequency energy ratio, $g(\alpha)$ is a double-sided opacity gate, and $c_i$ is a Gaussian band-pass on DC strength.\\

\noindent\textbf{Redundancy Features $z_3$.}
Based on $\mathcal{C}_{\text{red}}$, we characterize distributional density among 3D Gaussians and estimate substitutability via overlap-weighted neighborhood similarity in DC color and shape:
\begin{equation}
z_3(i)=\mathrm{Norm}\!\Big(\tfrac{1}{k}\sum_{j\in\mathcal{N}_k(i)} w_{ij}\,r_{ij}\Big),
\label{eq:z3}
\end{equation}
where $r_{ij}$ combines color and shape similarity, and $w_{ij}=\exp(-d_{ij}^2/(\sigma_o^2(\bar s_i^2+\bar s_j^2)))$ approximates projected overlap using spatial distance and scale.

Each expert $k$ maps its features $z_k(i)$ into an evidence package $E_k(i) = [U_k(i),S_k(i)]$, separating quality from certainty:
\begin{equation}
\begin{aligned}
U_k(i) &= \mathrm{Norm}\big(\mathrm{Disp}_{\mathcal{N}(i)}(z_k) + \mathrm{Penalty}_k(i)\big),\\
S_k(i) &= \mathrm{Norm}(z_k(i)), \ k\in\{1,2,3\}.
\end{aligned}
\label{eq:evidence_mapping}
\end{equation}
where $U_k\in[0,1]$ measures uncertainty from neighborhood dispersion ($\mathrm{Disp}$) and expert-specific penalties, and $S_k\in[0,1]$ is the quality score; this decoupling enables confidence-aware expert gating.


\subsection{Safety and Budget-Aware Gate (SBAG)}
\label{subsec:carrier_maker}
To select robust watermark carriers from a large 3D-GS set, we decouple the pipeline into \emph{ranking} and \emph{budgeting}.
Ranking relies solely on Trio-Experts evidence packages $E_k(i) = [U_k(i),S_k(i)]$, mapped to proxy scores aligned with
$\{\mathcal{C}_{\text{geo}},\mathcal{C}_{\text{app}},\mathcal{C}_{\text{red}}\}$:
\begin{equation}
R_k(i)=\mathrm{clip}\!\big(S_k(i)-\beta U_k(i),\,0,\,1\big),\quad k\in\{1,2,3\},
\label{eq:rk_clip}
\end{equation}
where $\beta$ is a fixed constant shared across experts. We interpret $R_1$ as geometric stability, $R_2$ as appearance safety, and $R_3$ as redundancy certainty; in all cases, higher scores are preferred.
To avoid assuming expert dominance, we define a symmetric point-wise utility:
\begin{equation}
u_i=\Big(R_1(i)\cdot R_2(i)\cdot R_3(i)\Big)^{\tfrac{1}{3}}.
\label{eq:ui_sym}
\end{equation}

We render all training views once using DC+opacity to obtain view-corrected visibility and distribution priors.
The rasterizer provides the non-negative compositing weight $w^{(t)}_{i,p}\ge 0$ of Gaussian $i$ at pixel $p$ in view $t$.
We estimate a scene-level crowding factor $\eta\in(0,1]$ as
\begin{equation}
\eta=\frac{1}{V}\sum_{t=1}^{V}
\frac{\sum_{p}\min\!\Big(1,\ \sum_i w^{(t)}_{i,p}\Big)}
{\sum_{p}\sum_i w^{(t)}_{i,p}+\epsilon},
\qquad w^{(t)}_{i,p}\ge 0,
\label{eq:eta_overlap}
\end{equation}
and obtain per-Gaussian visibility $v_i\in[0,1]$ by accumulating its screen-space contribution and normalizing.
We further compute the scene-average visibility $\bar v=\frac{1}{N}\sum_{i=1}^{N} v_i$.

Given message length $M$ (bits), we model the effective bits contributed by one carrier as a scene-adaptive coefficient
\begin{equation}
\kappa_{\mathrm{eff}}=\kappa_{0}\cdot \bar v \cdot \eta,\qquad
B=\left\lceil \frac{M}{\kappa_{\mathrm{eff}}}\right\rceil,
\label{eq:self_budget_B}
\end{equation}
where $\kappa_{0}$ is a constant determined by the embedding design.

We define a generic feasible set using quantile-based constraints:
\begin{equation}
\mathcal{F}=\Bigl\{\, i\ \Big|\ 
\begin{aligned}
& R_1(i)\ge \mathrm{Q}_{q}(R_1),\ 
R_2(i)\ge \mathrm{Q}_{q}(R_2),\\
& R_3(i)\ge \mathrm{Q}_{q}(R_3),\ 
v_i \ge \mathrm{Q}_{q}(v)
\end{aligned}
\Bigr\},
\label{eq:feasible_set_quantile}
\end{equation}
and perform deterministic water-level selection within $\mathcal{F}$ by $u_i$:
\begin{equation}
\mathcal{WM}_0=\text{top-}B\ \{\,u_i\mid i\in\mathcal{F}\,\}.
\label{eq:seed_topB}
\end{equation}

To bridge viewpoint-induced coverage gaps, we retain a \textbf{Prototype-based Proximity Extension}.
We build a compact evidence vector where $c_i=\|\mathbf{h}_i^{(0)}\|_2$ is DC strength and $h_i=\rho_i^{\mathrm{hf}}$ is the AC high-frequency ratio:
\begin{equation}
\mathbf{e}_i=\mathrm{Norm}\big(R_1(i),\, R_2(i),\, R_3(i),\,v_i,\,h_i,\,c_i\big),
\label{eq:evidence_vec}
\end{equation}
compute the prototype $\boldsymbol{\mu}=\frac{1}{|\mathcal{WM}_0|}\sum_{i\in\mathcal{WM}_0}\mathbf{e}_i$,
recruit proximal neighbors $\mathcal{WM}_{\mathrm{prox}}$ by cosine similarity, and form
\begin{equation}
\mathcal{WM}_{\mathrm{parent}}=\mathcal{WM}_0\cup \mathcal{WM}_{\mathrm{prox}}.
\label{eq:parent_set}
\end{equation}
Finally, each parent is split into $N_s$ visually equivalent children, routing one child to the watermark branch:
\begin{equation}
\mathcal{WM}_{\star}=\bigcup_{i\in\mathcal{WM}_{\mathrm{parent}}}\mathcal{C}_{\mathrm{wm}}(i),
\label{eq:wstar}
\end{equation}
while the remaining children act as visual compensators during finetuning to neutralize embedding artifacts.
The resulting $\mathcal{WM}_\star$ and its complement $\mathcal{VIS}$ are used for subsequent channel-wise mask routing.
\subsection{Channel-wise Group Mask}
To avoid visible degradation, we assign channel-wise masks to watermark carriers and visual compensators, and optimize a \emph{visual loss} and a \emph{watermark loss} under separate gradient routes. For Gaussian $i$, we precompute per-group masks $m^{\mathrm{wm}}_{g}(i), m^{\mathrm{vis}}_{g}(i)\in[0,1]$ from Trio evidence and one-shot priors, 
We derive the two masks from point-wise channel weights$w^{(\mathrm{vis})}_{g}(j)$ and $w^{(\mathrm{wm})}_{g}(j)$ as
\begin{equation}
\begin{aligned}
m^{\mathrm{vis}}_g &= \max \! \big( \mathrm{clip}( \mathrm{mean}(w^{\mathrm{vis}}_g[\mathcal{VIS}]),\, 0,\, cap_g ),\, floor_g \big),\\
m^{\mathrm{wm}}_g &= \mathrm{clip} \big( \mathrm{med}(w^{\mathrm{wm}}_g[\mathcal{WM}_\star]),\, 0,\, cap_g \big).
\end{aligned}
\label{eq:masks}
\end{equation}
where $Q_{0.5}$ is the median, $\mathcal{V}=\{1,\dots,N\}\setminus \mathcal{WM}_\star$, and $\mathrm{cap}_g,\mathrm{floor}_g$ are per-channel bounds.

and gate gradients as
\begin{equation}
\label{eq:gate}
\nabla_{\theta^g_i}\mathcal{L}=
\begin{cases}
m^{\mathrm{wm}}_{g}(i)\, \nabla_{\theta^g_i}\mathcal{L}_{\mathrm{wm}}, & i\in \mathcal{WM}_\star,\\[2pt]
m^{\mathrm{vis}}_{g}(i)\, \nabla_{\theta^g_i}\mathcal{L}_{\mathrm{vis}}, & i\in \mathcal{VIS},
\end{cases}
\end{equation}
where $g \in \{\boldsymbol{\delta}_{\mathrm{dc}}, \boldsymbol{\rho}_{\mathrm{rest}}, \boldsymbol{\omega}_{\mathrm{opa}}, \boldsymbol{\theta}_{\mathrm{rot}}, \boldsymbol{\sigma}_{\mathrm{sca}}\}$ denotes five parameter channels.

To ensure complete gradient separation, we perform two passes of $\mathcal{L}_{\mathrm{vis}}$ and $\mathcal{L}_{\mathrm{wm}}$, so that $\mathcal{WM}_\star$ and $\mathcal{VIS}$ receive their gradients in an orthogonal manner.

\begin{figure*}[h]
    \centering
    \includegraphics[width=\textwidth]{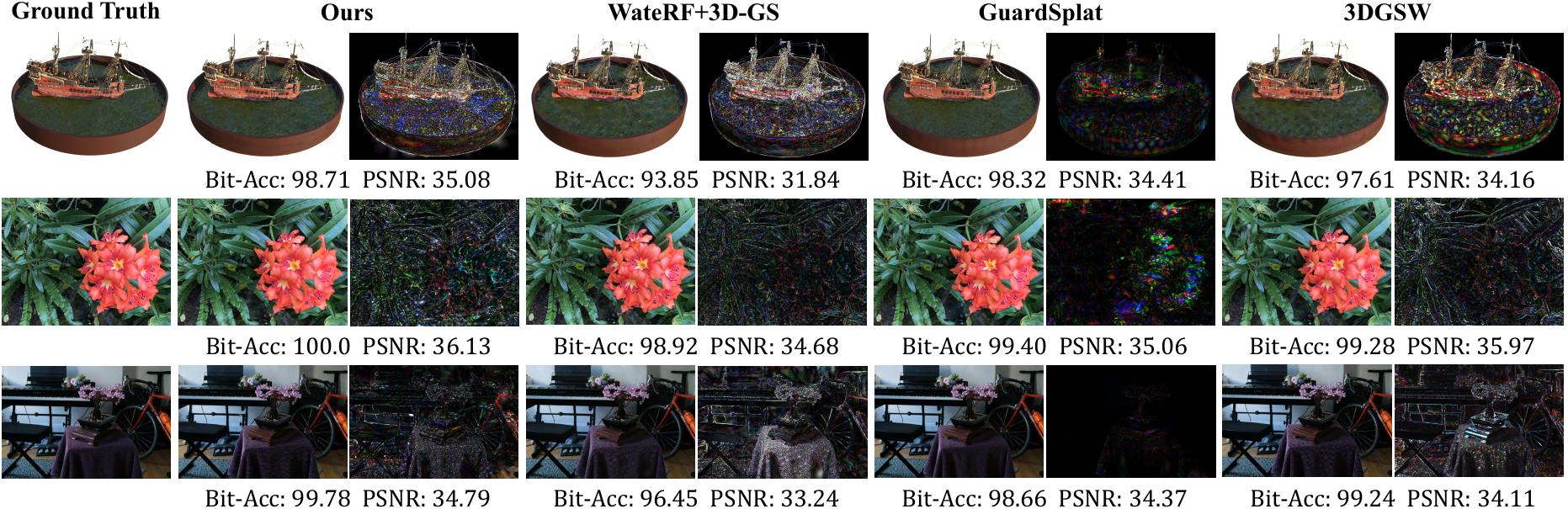}
    \vspace{-1.6em}
    \caption{\textbf{Rendering-quality comparison.} We compare our method with all baselines using 32-bit messages. The difference maps are shown at 10× scale. Our approach achieves higher bit accuracy and better visual fidelity than competing methods.}
    \label{fig:com}
\end{figure*}

\begin{table*}[htb!]
\setlength{\tabcolsep}{3pt}
\centering
\footnotesize{
\begin{tabular}{@{}l|ccccc|ccccc|ccccc@{}}
\toprule
  & \multicolumn{4}{c}{32 bits} & & \multicolumn{4}{c}{48 bits} & & \multicolumn{4}{c}{64 bits} \\
   \cmidrule{2-5} \cmidrule{7-10} \cmidrule{12-15} 
    Methods   & Bit Acc↑ & PSNR ↑ & SSIM ↑ & LPIPS ↓ & & Bit Acc↑ & PSNR ↑ & SSIM ↑ & LPIPS ↓ & & Bit Acc↑ & PSNR ↑ & SSIM ↑ & LPIPS ↓  \\ \midrule
     WateRF~\cite{waterf}+3D-GS~\cite{3D-GS} & 93.28 & 30.57 & 0.954 & 0.052 &&  84.39  &  30.06  & 0.949 & 0.056  &&  74.92 & 25.73 & 0.887 & 0.105 \\
     GuardSplat~\cite{guardsplat}  & 95.58 & 35.32 & 0.978 & 0.043 &&  93.29 & 33.36 & 0.969 & 0.045  &&  90.14 & 32.25 & 0.963 & 0.048 \\ 
      3D-GSW \cite{3dgsw}  & 97.22 & 35.15 & 0.977 & 0.044 && 93.59 & 33.26 & 0.972 & 0.047 && 91.31 & 32.52 & 0.966 & 0.050 \\
     \midrule
     Ours  & \textbf{98.46}   &  \textbf{35.98}  & \textbf{0.982} & \textbf{0.041}  &&  \textbf{94.29}  &  \textbf{33.45}  & \textbf{0.973} & \textbf{0.044} &&  \textbf{91.65} & \textbf{32.71} & \textbf{0.969} & \textbf{0.047} \\
\bottomrule
\end{tabular}}
\vspace{-1em}
\caption{\textbf{Bit accuracy and rendering quality compared with baselines.} We report results for 32, 48, and 64-bit messages, averaged over the Blender, LLFF, and Mip-NeRF 360 datasets. The best scores are shown in \textbf{bold}.}
\vspace{-1.4em}
\label{tab:invisibility_capacity}
\end{table*}

\subsection{Decoupled Watermark Finetuning}
\label{subsec:decoupled_finetuning}
To resolve the gradient conflict between rendering fidelity and watermark robustness, we implement a \textbf{decoupled finetuning} strategy.
It isolates gradient propagation by ensuring that watermark carriers $\mathcal{WM}_\star$ are exempt from visual reconstruction constraints, while visual compensators $\mathcal{VIS}$ remain unaffected by watermark-related objectives.
Specifically, the visual objective is
\begin{equation}
\mathcal{L}_{\mathrm{vis}}
= \lambda_{\mathrm{rec}}\mathcal{L}_{\mathrm{rec}}
+ \lambda_{\mathrm{lpips}}\mathcal{L}_{\mathrm{lpips}}
+ \lambda_{\mathrm{wav}}^{\mathrm{high}}\,\mathcal{L}_{\mathrm{wav}}^{\mathrm{high}},
\label{eq:Lvis}
\end{equation}
where $\mathcal{L}_{\mathrm{rec}}$ is the L1 reconstruction loss, $\mathcal{L}_{\mathrm{lpips}}$ is LPIPS~\cite{lpips}, and
$\mathcal{L}_{\mathrm{wav}}^{\mathrm{high}}$ penalizes multi-level DWT high-frequency subbands ($S\in\{LH,HL,HH\}$) with an L1 distance.

Simultaneously, we optimize the watermark branch using Expectation Over Transformation (EOT).
Let $\mathbf{b}_i\in\{0,1\}^B$ denote the target bits of carrier $i$, $\mathcal{V}_i$ its visible views, and $D_\psi(\cdot)$ a decoder producing logits.
For clean and transformed renderings, we average logits over visible views:
\begin{equation}
\bar{\mathbf{z}}_i^{\mathrm{cl}}=\tfrac{1}{|\mathcal{V}_i|}\sum_{v\in\mathcal{V}_i} D_\psi(\hat{I}_v), \quad
\bar{\mathbf{z}}_i(\mathsf{t})=\tfrac{1}{|\mathcal{V}_i|}\sum_{v\in\mathcal{V}_i} D_\psi(\mathsf{t}(\hat{I}_v)),
\label{eq:avglogits}
\end{equation}
where $\mathsf{t}\sim p(\mathsf{t})$ is sampled from a standard degradation family (e.g., blur, rotation, scaling, crop, noise, JPEG).
We decode bits by $\hat{\mathbf{y}}_i^{\mathrm{cl}}=\sigma(\bar{\mathbf{z}}_i^{\mathrm{cl}})$ and
$\hat{\mathbf{y}}_i^{\mathrm{eot}}(\mathsf{t})=\sigma(\bar{\mathbf{z}}_i(\mathsf{t}))$.
We embed watermarks \emph{only} in the DWT low-frequency ($LL$) subband and regularize low-frequency distortion with
$\mathcal{L}_{\mathrm{wav}}^{\mathrm{low}}$ (multi-level DWT $LL$ subband with an L1 distance; see supp.).
The clean and EOT watermark losses are
\begin{equation}
\begin{aligned}
\mathcal{L}_{\mathrm{wm}}^{\mathrm{clean}}
&=\tfrac{1}{|\mathcal{W}_\star|}\sum_{i\in\mathcal{W}_\star}\mathrm{BCE}\!\big(\hat{\mathbf{y}}_i^{\mathrm{cl}},\,\mathbf{b}_i\big),\\
\mathcal{L}_{\mathrm{wm}}^{\mathrm{eot}}
&=\mathbb{E}_{\mathsf{t}\sim p}\!\left[\tfrac{1}{|\mathcal{W}_\star|}\sum_{i\in\mathcal{W}_\star}
\mathrm{BCE}\!\big(\hat{\mathbf{y}}_i^{\mathrm{eot}}(\mathsf{t}),\,\mathbf{b}_i\big)\right],
\label{eq:Lwm-eot} 
\end{aligned}
\end{equation}
and the total watermark objective is
\begin{equation}
\mathcal{L}_{\mathrm{wm}}
= \lambda_{\mathrm{wm}}^{\mathrm{clean}}\,\mathcal{L}_{\mathrm{wm}}^{\mathrm{clean}}
+ \lambda_{\mathrm{wm}}^{\mathrm{eot}}\,\mathcal{L}_{\mathrm{wm}}^{\mathrm{eot}}
+ \lambda_{\mathrm{wav}}^{\mathrm{low}}\,\mathcal{L}_{\mathrm{wav}}^{\mathrm{low}}.
\label{eq:Lwm}
\end{equation}

Finally, although we report the joint scalar objective
$\mathcal{L}_{\mathrm{total}}=\lambda_{\mathrm{vis}}\mathcal{L}_{\mathrm{vis}}+\lambda_{\mathrm{wm}}\mathcal{L}_{\mathrm{wm}}$,
the actual parameter updates follow the masked routing rule in Eq.~\eqref{eq:gate}:
watermark gradients act only on $\mathcal{W}_\star$, and visual gradients act only on $\mathcal{VIS}$.
This disjoint gradient routing effectively eliminates the optimization interference in joint finetuning.

%% file: sec/4_experiment.tex
\section{Experiments}
\label{sec:experiment}
\subsection{Experimental Setting}

\textbf{Dataset \& Pre-trained 3D-GS.} We evaluate on the standard benchmarks in the NeRF \cite{nerf} and 3D-GS \cite{3D-GS} literature—Blender \cite{nerf2020eccv}, LLFF \cite{llff}, and Mip-NeRF 360 \cite{mipnerf}. Following common practice, we report results on 25 scenes drawn from the full versions of these datasets.

\begin{figure*}[t]
    \centering
    \includegraphics[width=\textwidth]{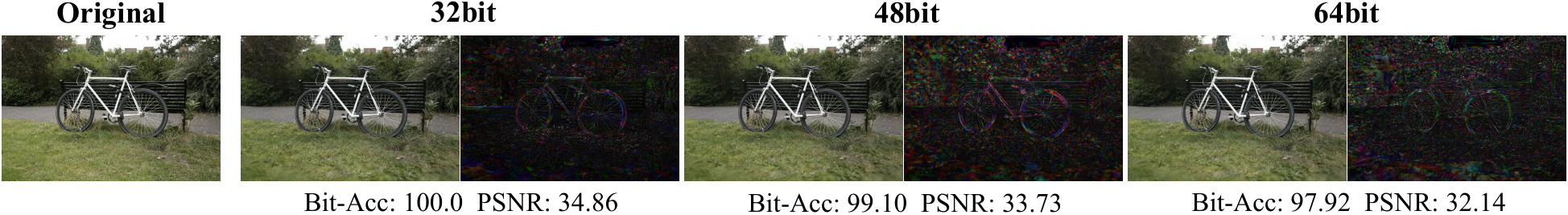}
    \vspace{-1.6em}
    \caption{\textbf{Rendering-quality comparison under different message capacities.} With 32-bit, 48-bit, and 64-bit embedded messages, (differences $\times 10$), our method maintains high bit-acc across different capacities while preserving perceptual quality.}
    \label{fig:cap}
    \vspace{-1.0em}
\end{figure*}

\begin{table*}[h]
\setlength{\tabcolsep}{3pt}
\begin{adjustbox}{max width=\textwidth,center}
{\large
\begin{tabular}{l|ccccccccc}
\toprule
\multicolumn{1}{c|}{} & \multicolumn{8}{c}{Bit Accuracy(\%) $\uparrow$} \\
\cmidrule{2-9}
\multicolumn{1}{c|}{Methods} & No Distortion & \begin{tabular}[c]{@{}c@{}}Gaussian Noise\\ ($\sigma$ = 0.1)\end{tabular} & \begin{tabular}[c]{c}Rotation\\ ($\pm \pi/6$)\end{tabular} & \begin{tabular}[c]{c}Scaling\\ (75\%)\end{tabular} & \begin{tabular}[c]{c}Gaussian Blur\\ ($\sigma$ = 0.1)\end{tabular} & \begin{tabular}[c]{@{}c@{}}Crop\\ (40\%)\end{tabular}  & \begin{tabular}[c]{@{}c@{}}JPEG Compression\\ (50\% quality)\end{tabular} &\begin{tabular}[c]{c}Combined\end{tabular} \\ 
\midrule
WateRF~\cite{waterf}+3D-GS~\cite{3D-GS} & 93.28 & 78.12 & 81.47 & 84.63 & 87.09 & 84.58 & 82.03 & 64.73 \\ 
GuardSplat~\cite{guardsplat} & 95.58 & 90.11 & 95.87 & 94.93 & 97.16 & 95.05 & 89.92 & 88.64 \\
3D-GSW~\cite{3dgsw} & 97.22 & 83.71 & 88.05 & 94.58 & 95.94 & 92.73 & 92.54 & 90.96  \\ \midrule
Ours &  \textbf{98.46}  & \textbf{91.22} & \textbf{96.18} & \textbf{95.06} &  \textbf{97.75} & \textbf{95.88} & \textbf{92.95} & \textbf{91.30} \\
\bottomrule
\end{tabular}
}
\end{adjustbox}
\vspace{-.7em}
\caption{\textbf{Quantitative robustness comparison under different attacks against baseline methods.} Results are averaged over the Blender, LLFF, and Mip-NeRF 360 datasets using 32-bit messages. The best scores are marked in \textbf{bold}.}
\label{tab:robustness}
\end{table*}

\noindent\textbf{Baseline.} We compare our method against three strategies for fairness: (i) WateRF~\cite{waterf}, an innovative watermarking method deployable in both the NeRF representations; for fairness, we instantiate and evaluate it in the 3D-GS setting. (ii) GuardSplat~\cite{guardsplat}, a CLIP-guided and SH-aware robust watermarking model for 3DGS, enabling invisible, secure, and distortion-resistant embedding with high visual fidelity. (iii) 3D-GSW~\cite{3dgsw}, a state-of-the-art watermark embedding model operating in the frequency domain.

\noindent\textbf{Implementation Details.}
We train all models on a single NVIDIA A800 GPU for 2–10 epochs using Adam~\cite{adam}. The decoder is a frozen HiDDeN~\cite{hidden} model (32/48/64 bits). We prune Gaussians with $V_{\pi}{<}10^{-8}$. All experiments are repeated with three random seeds.

\noindent\textbf{Evaluation.} We evaluate our watermarking framework along three key dimensions.\textbf{(1) Invisibility.}
Visual fidelity is measured using Peak Signal-to-Noise Ratio (PSNR), Structural Similarity Index (SSIM) \cite{ssim}, and Learned Perceptual Image Patch Similarity (LPIPS)~\cite{lpips}. \textbf{(2) Robustness.} We assess robustness by reporting bit accuracy under a variety of image-space degradations: Gaussian noise ($\sigma=0.1$), rotation (randomly sampled from $\pm \pi/6$), scaling to $75\%$ of the original size, Gaussian blur ($\sigma=0.1$), cropping to $40\%$, JPEG compression (quality $50\%$), and a composite attack that combines Gaussian noise, crop, and JPEG compression.  We also evaluate model-space distortions by removing 20\% of Gaussians, cloning 20\% of Gaussians, and adding Gaussian noise (\(\sigma=0.1\)) to 3D-GS parameters. \textbf{(3) Capacity.} We further investigate bit accuracy under different payload sizes, with message lengths $M_b \in \{32, 48, 64\}$.

\subsection{Results}
\noindent\textbf{Rendering Quality and Bit Accuracy.} We compare our approach with existing methods in terms of visual fidelity and watermark extraction. As shown in Fig.~\ref{fig:com}, our results are closest to the original views while still achieving high bit accuracy. This is especially notable on real-world scenes, where complex geometry makes faithful rendering harder. Unlike prior methods that struggle to trade off image quality and watermark reliability, our approach maintains both. Quantitative results in Tab.~\ref{tab:invisibility_capacity} further confirm that our method consistently preserves rendering quality and bit accuracy across all datasets.

\noindent\textbf{Robustness to Image Distortions.} In this section, we evaluate how well our method withstands post-processing on rendered images, where such operations may alter or weaken the embedded watermark. We measure bit accuracy under various distortion types. As shown in Table.~\ref{tab:robustness}, most existing methods fail to maintain high robustness. Moreover, WateRF+3D-GS and 3D-GSW without EOT-based adversarial training still struggle to reliably preserve embedded messages in rendered views. GuardSplat does introduce EOT to improve robustness, but its heavy reliance on the CLIP decoder and SH-space embedding makes it less effective under more complex or compound distortions. In contrast, our method performs strict selection and decoupled embedding, writing watermarks only into safe 3D Gaussians, thereby achieving consistently robust extraction across all distortion settings.

\begin{table}[t!]
\setlength{\tabcolsep}{1pt}
\begin{adjustbox}{max width=\textwidth,center}
\centering
\scriptsize{
\begin{tabular}{@{}l|cccc@{}}
\toprule
&  \multicolumn{3}{c}{Bit Accuracy(\%) $\uparrow$}  \\
    \cmidrule{2-5}
    Methods & \begin{tabular}[c]{@{}c@{}}No \\ Distortion \end{tabular} & \begin{tabular}[c]{@{}c@{}} Adding \\ Gaussian Noise \\ ($\sigma$ = 0.1) \end{tabular} & \begin{tabular}[c]{@{}c@{}} Removing \\ 3D Gaussians \\ (20 \%) \end{tabular} & \begin{tabular}[c]{@{}c@{}} Cloning \\ 3D Gaussians \\ (20 \%) \end{tabular} \\ \midrule
WateRF~\cite{waterf}+3D-GS~\cite{3D-GS} & 93.28 & 62.35 & 60.91 & 76.16 \\
GuardSplat~\cite{guardsplat}            & 95.58 & 79.96 & 88.49 & 92.30 \\ 
3D-GSW~\cite{3dgsw}                     & 97.22 & 89.93 & 97.23 & 96.85 \\ \midrule
Ours                                    & \textbf{98.46} & \textbf{90.52} & \textbf{98.16} & \textbf{97.48} \\

\bottomrule
\end{tabular}}
\end{adjustbox}
\vspace{-0.5em}
\caption{Robustness under model-level distortions with a 32-bit payload. Best results are shown in \textbf{bold}.}
\label{tab:robustness_3d}
\end{table}

\noindent\textbf{Robustness to Model-level Distortions.} To evaluate robustness under malicious model-level tampering, we perturb the 3D-GS representation by adding parameter noise and randomly removing or cloning 3D Gaussians. As reported in Tab.~\ref{tab:robustness_3d}, our method consistently outperforms prior works across these distortions, indicating that the embedded message is not tied to a fragile subset of primitives. Notably, it maintains reliable decoding even when the underlying model is partially corrupted, demonstrating robust protection for both the 3D-GS model and its rendered outputs.

\noindent\textbf{Message Capacity.} Since there is an inherent trade-off among bit accuracy, rendering quality, and payload size, we investigate different message lengths {32, 48, 64}. As shown in Table.~\ref{tab:invisibility_capacity}, increasing the message length consistently leads to a slight drop in both bit accuracy and rendering quality. Nevertheless, our method maintains a favorable balance between invisibility and capacity, and the performance gap over other methods becomes more evident as the message length grows. Fig.~\ref{fig:cap} further illustrates that our approach achieves a better compromise between bit accuracy and visual fidelity.

\subsection{Ablation Study}
\begin{figure}[h]
    \centering
    \includegraphics[width=\linewidth]{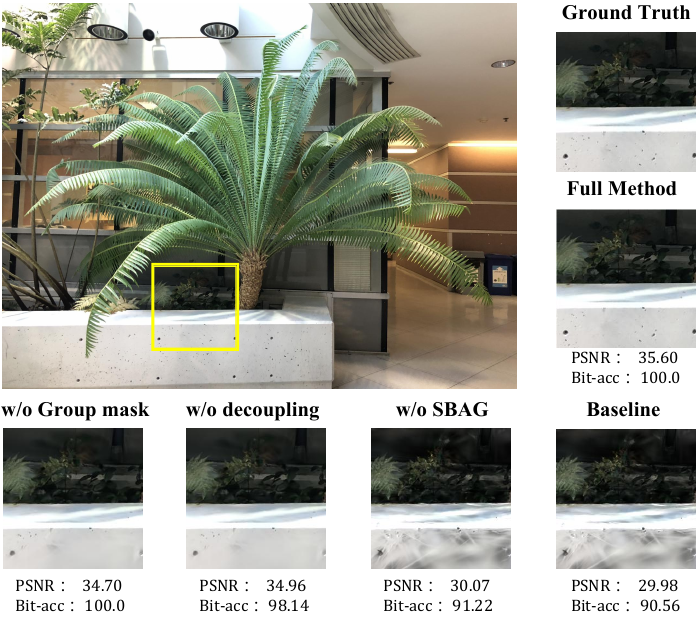}
    \vspace{-1.6em}
    \caption{Comparison of rendering quality among the full method (ours), without SBAG, without group mask, without decoupled finetuning, and the baseline model. All images are embedded with 32-bit messages.}
    \label{fig:ab_method}
    \vspace{-1.0em}
\end{figure}

In this section, we ablate the \textbf{SBAG}, \textbf{Group Mask}, and \textbf{Decoupled Finetuning} to clarify their necessity. Fig.~\ref{fig:ab_method} and Tab.~\ref{tab:ablation_method} show that removing any one of them leads to either lower bit accuracy or degraded rendering, and removing all of them breaks the quality–robustness balance. This indicates that all three are required to stabilize watermark embedding.

\begin{table}[h]
\centering
\scriptsize
\resizebox{\columnwidth}{!}{
\begin{tabular}{@{}ccc|cccc@{}}
\toprule
\multicolumn{3}{c}{Methods} & \multicolumn{4}{c}{Ours} \\
\cmidrule{1-7}
SBAG & Group Mask & Decoupling & Bit Acc(\%)↑ & PSNR↑ & SSIM↑ & LPIPS↓ \\ \midrule
--            & --            & --            & 94.70 & 30.00 & 0.952 & 0.073 \\
\checkmark    & \checkmark    & --            & 96.80 & 34.10 & 0.974 & 0.048 \\
\checkmark    & --            & \checkmark    & 97.10 & 33.70 & 0.972 & 0.051 \\
--            & \checkmark    & \checkmark    & 95.10 & 33.10 & 0.968 & 0.056 \\
\checkmark    & \checkmark    & \checkmark    & \textbf{97.80} & \textbf{35.20} & \textbf{0.979} & \textbf{0.042} \\
\bottomrule
\end{tabular}
}
\caption{\textbf{Quantitative ablation results.} The best performance is obtained when all components are activated, with evaluation conducted on 32-bit message embedding.}
\label{tab:ablation_method}
\end{table}

\noindent\textbf{SBAG.} To jointly improve bit accuracy and rendering quality during watermark embedding, we introduce the SBAG to select, expand and create high-quality carrier Gaussians. The Gaussians selected by SBAG lie in visually insensitive regions and have multiple channels that can be safely modified. Combined with decoupled training, this effectively alleviates the conflict between visual fidelity and watermark embedding. As shown in Tab.~\ref{tab:ablation_method} and Fig.~\ref{fig:ab_method}, SBAG-based selection consistently boosts both rendering quality and bit accuracy. Moreover, Tab.~\ref{tab:DWT_subband_loss} shows that controlled densification within an adaptive budget can further improve watermark performance with lower storage cost.\\

\begin{table}[t!]
\centering
\setlength{\tabcolsep}{9pt}
\scriptsize{
\begin{tabular}{@{}c|ccccc@{}}
\toprule
Budget& Bit Acc ↑ & PSNR ↑ & SSIM ↑ & LPIPS ↓ & Storage ↓ \\ \midrule
$1\%$  & 97.19 & 34.98 & 0.980 & 0.046 & \textbf{164MB} \\
Adaptive  & \textbf{98.46} & \textbf{35.98} & \textbf{0.982} & \textbf{0.041} & 178MB \\
$10\%$ & 98.12 & 34.24 & 0.975 & 0.048 & 195MB \\ 
\bottomrule
\end{tabular}
}
\vspace{-1.2em}
\caption{\label{tab:DWT_subband_loss} Ablation study on Budget for Watermark Carrier Gaussians. Results represent the average score across Blender, LLFF, and Mip-NeRF 360 datasets using 32-bit messages.}
\vspace{-1.4em}
\end{table}

\begin{figure}[h]
    \centering
    \includegraphics[width=\linewidth]{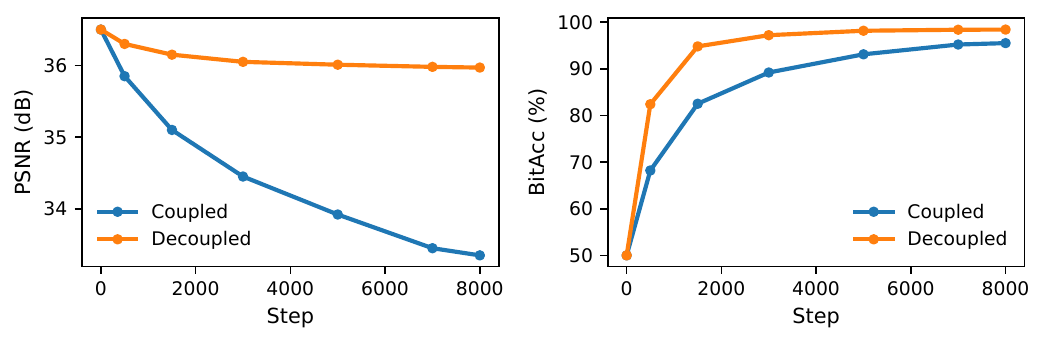}
    \vspace{-1.6em}
    \caption{\textbf{Training dynamics of coupled vs. decoupled finetuning.} Decoupled optimization converges faster and consistently achieves higher PSNR and BitAcc across finetuning steps.}
    \label{fig:ab_decouple}
\end{figure}

\begin{figure}[t]
    \centering
    \includegraphics[width=\linewidth]{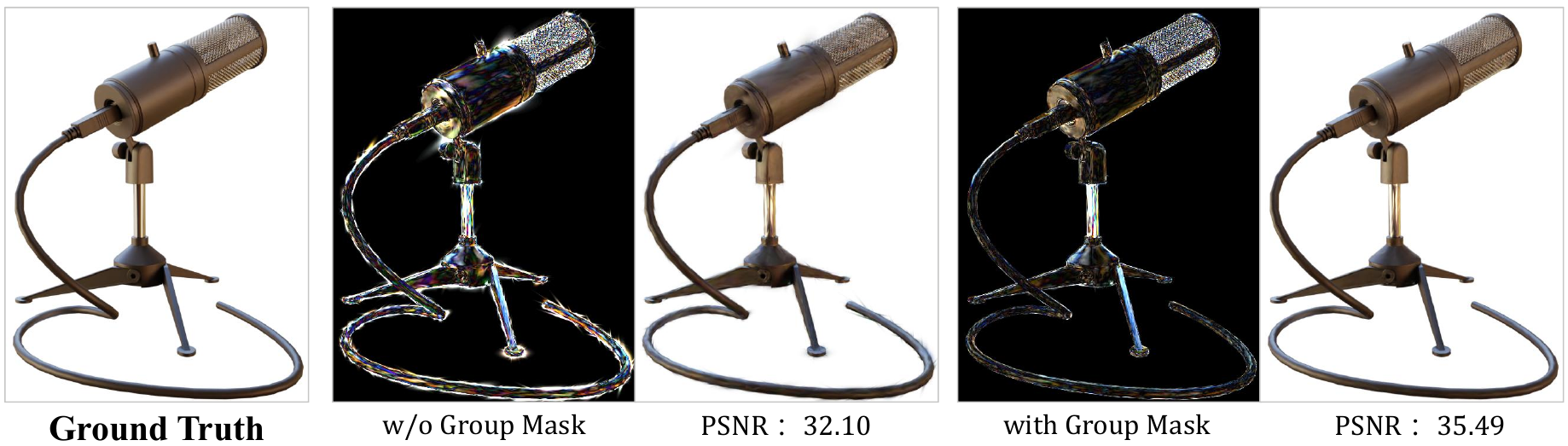}
    \vspace{-1.2em}
    \caption{\textbf{Qualitative comparison of our group mask.} For background-free objects, it confines 3D Gaussian updates to object boundaries, preserving rendering quality.(difference $\times 8$)}
    \label{fig:ab_mask}
    \vspace{-1.2em}
\end{figure}

\noindent\textbf{Group mask.} Before watermark embedding, the pretrained 3D-GS already achieves high rendering quality, so we want to avoid large updates on the majority of non-carrier (i.e., visual) Gaussians. In contrast, carrier Gaussians should have more freedom to be perturbed across channels to encode the watermark. Our gradient group mask enforces this asymmetry during backpropagation: gradients on visual Gaussians are suppressed below those of prior methods, while gradients on carrier Gaussians are adaptively scaled per channel. This preserves rendering quality while maintaining watermark accuracy. As shown in Fig.~\ref{fig:ab_mask}, our group-aware mask achieves a better balance between visual fidelity and watermark performance than alternative masking strategies.\\

\noindent\textbf{Decoupled Finetuning.} To verify the necessity of decoupled finetuning, we analyze the optimization conflict between the visual and watermark objectives by running coupled and decoupled finetuning under identical settings and periodically evaluating PSNR and BitAcc on the same validation set. As shown in Fig.~\ref{fig:ab_decouple}, coupled finetuning exhibits a transient degradation in BitAcc while PSNR keeps improving, revealing a clear trade-off, whereas decoupled optimization converges faster and maintains consistently higher PSNR and BitAcc throughout finetuning.\\

We provide additional ablation results in the supplementary material.

%% file: sec/5_conclusion.tex
\section{Conclusion}
\label{sec:conclusion}
We conclude with a representation-native framework that unifies where to write, what to write, and why it matters.
\textbf{Where:} Trio-Experts derive geometry, appearance, and redundancy priors from Gaussian parameters, and together with a safety and budget aware SBAG plus single-pass rendering cues, select carriers that are robust yet visually safe, training them separately from visual Gaussians.
\textbf{What:} A channel wise group mask constrains gradients to designated parameter channels, preserving rendering quality while encoding the signal.
\textbf{Why:} This separation makes watermarking controllable and interpretable, sustains high bit accuracy under image-space and model-space attacks, and yields per Gaussian attribution for auditable verification; the design naturally extends to dynamic scenes and multimodal payloads.
\vspace{-1.6em}
\paragraph{Limitations.}
Our frequency decoupled training requires careful tuning of loss weights to balance quality and robustness; we provide empirically validated defaults, but extreme configurations can degrade performance. In addition, as with other neural watermarking methods, the approach depends on a pretrained decoder, and its robustness bounds the overall system performance.

\vspace{-0.3em}
\section*{Acknowledgements}
\vspace{-0.5em}

{\small
\noindent\raggedright
This work was supported by JST BOOST (Japan), Grant Number ~\mbox{JPMJBS2429}.\par
}